# Distance-to-Mean Continuous Conditional Random Fields to Enhance Prediction Problem in Traffic Flow Data

Sumarsih Condroayu Purbarani
Faculty of Computer Science
Universitas Indonesia
Indonesia
sumarsih.condroayu@ui.ac.id

Hadaiq Rolis Sanabila
Faculty of Computer Science
Universitas Indonesia
Indonesia
hadaiq@cs.ui.ac.id

Wisnu Jatmiko
Faculty of Computer Science
Universitas Indonesia
Indonesia
wisnuj@cs.ui.ac.id

*Abstract*— The increase of vehicle in highways may cause traffic congestion as well as in the normal roadways. Predicting the traffic flow in highways especially, is demanded to solve this congestion problem. Predictions on time-series multivariate data, such as in the traffic flow dataset, have been largely accomplished through various approaches. The approach with conventional prediction algorithms, such as with Support Vector Machine (SVM), is only capable of accommodating predictions that are independent in each time unit. Hence, the sequential relationships in this time series data is hardly explored. Continuous Conditional Random Field (CCRF) is one of Probabilistic Graphical Model (PGM) algorithms which can accommodate this problem. The neighboring aspects of sequential data such as in the time series data can be expressed by CCRF so that its predictions are more reliable. In this article, a novel approach called DM-CCRF is adopted by modifying the CCRF prediction algorithm to strengthen the probability of the predictions made by the baseline regressor. The result shows that DM-CCRF is superior in performance compared to CCRF. This is validated by the error decrease of the baseline up to 9% significance. This is twice the standard CCRF performance which can only decrease baseline error by 4.582% at most.

## I. Introduction

The pace of development accompanied by the growth rate in the number of vehicles inevitably has impact on increasing air pollution levels of the region. Highways are suggested as a solution to at least decrease the emission by pollutant [1]. Thus, highway's role is big enough to save the air quality. Moreover, highways allow one of pollutant contributors, motor vehicles, to move from a point to another in a shorter time compared to when using normal roadways. Hence, become an alternative for people with vehicle to use highway instead. The ease provided by highways brings about more vehicles use this public facility. The increase of vehicle in highways may cause traffic congestion as well. If it happens, then it is sign for road developer or policy makers to build the infrastructure to expand the highways.

Predicting the traffic flow in highways especially, is essential. Prediction result might be used by people to plan their journey accordingly. Thus, decreases the number of vehicles getting stuck in the highway traffic. Predicting the traffic flow could also contribute in the fastest route planning in emergency evacuation, like ambulance and firefighter cars movement.

The highway traffic data is a time series data with many predictor variables [2]. Continuous Conditional Random Field (CCRF) is one of several variants of the Probabilistic Graphical Model (PGM) that can accommodate multivariate time series prediction problems. Therefore, in this article, different approaches are proposed in exploring possible information from the interaction amongst nodes in CCRF graph. It aims to reinforce the probability of the prediction made by the baseline regressor.

Several researches on the implementation of CCRF in time series data has been done. Some of which are studies conducted by Radosavljevic et al. [3] which proposes a standard CCRF modification to solve a prediction problem in time series data. They implement CCRF in Aerosol Optical Depth (AOD) data by using two prediction results, i.e. statistical modeling and deterministic method. To increase CCRF's expression they also modify the edge feature by rewarding features that meet certain spatial criteria and penalize otherwise. This modification can well capture the spatial information from AOD data.

Another work presented in [4] uses CCRF to model emotion prediction based on face expression and audio. This work done by Baltrusaitis, et al. is even elaborated to the modified version on the CCRF's vertex or variable feature. The modification involves the use of neural network to make a baseline prediction then add up the values of neighboring nodes instead of taking the difference between them. This approach is called Continuous Conditional Neural Field (CCNF) [5]–[7].

## II. Methods

### A. Standard CCRF

Continuous Conditional Random Field (CCRF) is one of several Probabilistic Graphical Model (PGM) algorithms that can accommodate sequential prediction problems with many variables. CCRF was first introduced by Qin et al. In his research on global ranking problems in document classification [8]. CCRF is a regression version of CRF used for classification problems. CCRF can model conditional probabilities (in Probabilistic Density Function/PDF form) of predictive values based on predictive values by baseline regressor(s).

Sponsor: Directorate of Research and Community Service Universitas Indonesia under PITTA Grant

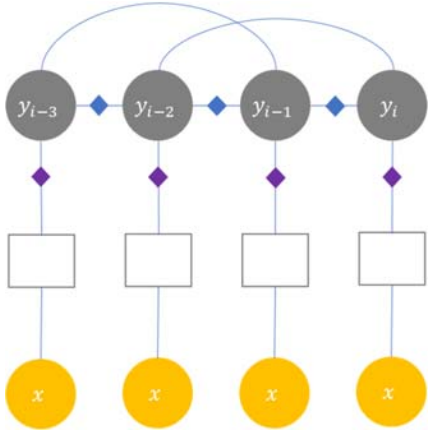

Fig. 1. CCRF structure

Similar with other PGM algorithms' working principles, CCRF establishes connections between a node and its neighboring nodes. These nodes represent prediction values at each time unit generated by conventional predictor algorithms as its baseline. The baseline regressor can be any regressors such as SVM, neural network algorithms or tree. Thus, CCRF serves to reinforce the probability of a weak prediction value. Figure 1 shows CCRF work scheme.

In general form, CCRF equation can be formulated as a PDF distribution as the following (1).

$$P(y|X) = \frac{\exp(\Psi)}{\int_y \exp(\Psi)} \quad (1)$$

where $y$ is predicted value and $X$ is random variable vector or predictor vector. $\Psi$ is the potential function of CCRF and is defined as (2).

$$\Psi(y, X, \alpha, \beta) = \sum_{i=1}^{N} F(\alpha, y_i, X) + \sum_i G(\beta, y_i, X) \quad (2)$$

with $F$ is CCRF's variable feature while $G$ is its edge feature. These two features are two sources of information used in CCRF. Variable feature contributes an *a priori* knowledge for CCRF since it evaluates the prediction results made by the baseline regressor. The edge feature $G$ shows interaction between predicted values. $F$ and $G$ can be further defined as follows (3)(4).

$$\sum_{i=1}^{N} F(\alpha, y_i, X) = -\sum_{i=1}^{N} \sum_{k=1}^{K1} \alpha_k (y_i - f_k(X_i))^2 \quad (2)$$

$$\sum_i G(\beta, y_i, X) = -\sum_i \sum_{k=1}^{K2} \beta_k (y_i - y_j)^2 \quad (3)$$

where $N$ represents the number of samples of observation. $K1, K2$ is the number of baseline regressors and the number of similarity between adjacent nodes ($i$-th and $j$-th nodes) respectively. $\alpha, \beta$ is contribution parameter of variable feature and edge feature respectively. $f_k(X_i)$ is the prediction by baseline regressor.

### B. DM-CCRF

In this research, the edge feature is modified to improve the performance of CCRF in predicting the time series data. The modification on the edge feature assumes that by knowing the average occurrence of an event in a sequence of events, the belief that the event will appear in the future is expected to increase. Hence, this novel approach is called Distance-to-Mean CCRF (DM-CCRF) and shown as Figure 2.

This assumption is formulated by introducing the new edge feature $H$ as follows (4).

$$\sum_i H(\theta, y_i, X) = -\sum_i \sum_{k=1}^{K3} \theta_k (y_i - m_i)^2 \quad (4)$$

where $K3$ is the sequence length being computed, $\theta$ is contribution parameter of the modified edge feature, and $m_i$ is defined as the following (5).

$$m_i = \frac{1}{i-1} \sum_{s=1}^{i-1} y_s \quad (5)$$

Thus, in the probabilistic form, DM-CCRF can be written as follows (6).

$$P(y|X) = \frac{1}{\eta} exp\left(-\sum_{i=1}^{N}\sum_{k=1}^{K1} \alpha_k (y_i - f_k(X_i))^2 - \sum_i \sum_{k=1}^{K3} \theta_k (y_i - m_i)^2\right) \quad (6)$$

where $\eta$ is the normalizer to keep the probability $P(y|X)$ value in between 0 and 1 and given as follows (7).

$$\eta = \int_y \exp(\Psi) \quad (6)$$

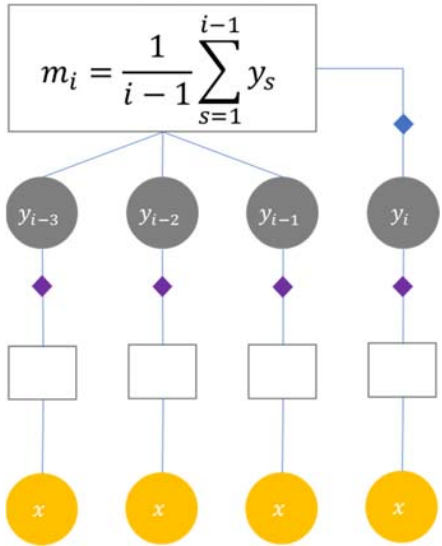

Fig. 2. DM-CCRF structure

In the form of computationally tractable matrix calculation, (6) can be written as follows (8).

$$P(y|X) = \frac{exp\left(-\frac{1}{2}(y-\mu)^T \sigma^{-1}(y-\mu)\right)}{(2\pi)^{n/2}|\sigma|^{1/2}} \quad (8)$$

where $\sigma^{-1}$ contains contribution parameters of DM-CCRF's overall feature function. $\mu$ represents the mean of predictor variables. $\sigma^{-1}$ and $\mu$ is formulated as (9) and (10).

$$\sigma^{-1} = A + C \quad (9)$$

$$\mu(X) = \sigma\tau \quad (10)$$

with $A, C, \tau$ is given as (11), (12), and (13) respectively.

$$A_{i,j} = \begin{cases} \sum_{k=1}^{K1} \alpha_k, & i = j \\ 0, & i \neq j \end{cases} \quad (11)$$

$$C_{i,j} = \begin{cases} \theta_k U_1, & i = j = 1 \\ \theta_k(1 + U_i), & i = j \in \{2, \dots, K3 - 1\} \\ -2\,\theta_k\left(\frac{1}{j-1} + U_j\right), & i \neq j \\ \theta_k, & i = j = K3 \end{cases} \quad (12)$$

$$\tau = 2 \sum_{k=1}^{K1} \alpha_k X_{i,j} \quad (13)$$

*C. Learning and Inference in DM-CCRF*

Learning in DM-CCRF aims to optimize $\alpha$ and $\theta$ parameter such that the probability value of $P(y|X)$ is maximized. This learning process can be written as (14).

$$(\alpha^*, \theta^*) = \underset{\alpha,\theta}{\mathrm{argmax}}\big(\log(P(y|X))\big) \quad (14)$$

Inference in DM-CCRF is objected to predict the unseen data. This can be done by predicting the most optimal $y$ such that can maximize the probability $P(y|X)$. This is just the same as seeking for the mean value of the unseen data's random variable distribution. Thus, it can be written as (15).

$$\hat{y} = \underset{y}{\mathrm{argmax}}\big(P(y|X)\big) = \mu(X) \quad (15)$$

## III. EXPERIMENTAL SETUP

*A. Dataset*

The experiment conducted in this work is implemented in Highways Agency dataset. It is a vehicle density data on a highway in the United Kingdom 2009-2013. It is a large dataset of 270,000,000 observations. Due to computational efficiency consideration, the observation coverage is limited by the latitude of 50.832657 hence left only 2,760 observations to be used in this study.

TABLE I. PARAMETER VARIATION OF ELM AS BASELINE REGRESSOR

| Scenario | Kernel Parameter | Regularization Coefficient |
|---|---|---|
| 1 | 1 | 1 |
| 2 | 1 | 5 |
| 3 | 1 | 10 |
| 4 | 1 | 50 |
| 5 | 1 | 100 |
| 6 | 1 | 500 |
| 7 | 1 | 1000 |
| 8 | 1 | 10000 |
| 9 | 1 | 1000000 |
| 10 | 1000000 | 5 |
| 11 | 1000000 | 10 |
| 12 | 1000000 | 50 |
| 13 | 1000000 | 100 |
| 14 | 1000000 | 1000 |
| 15 | 1000000 | 10000 |

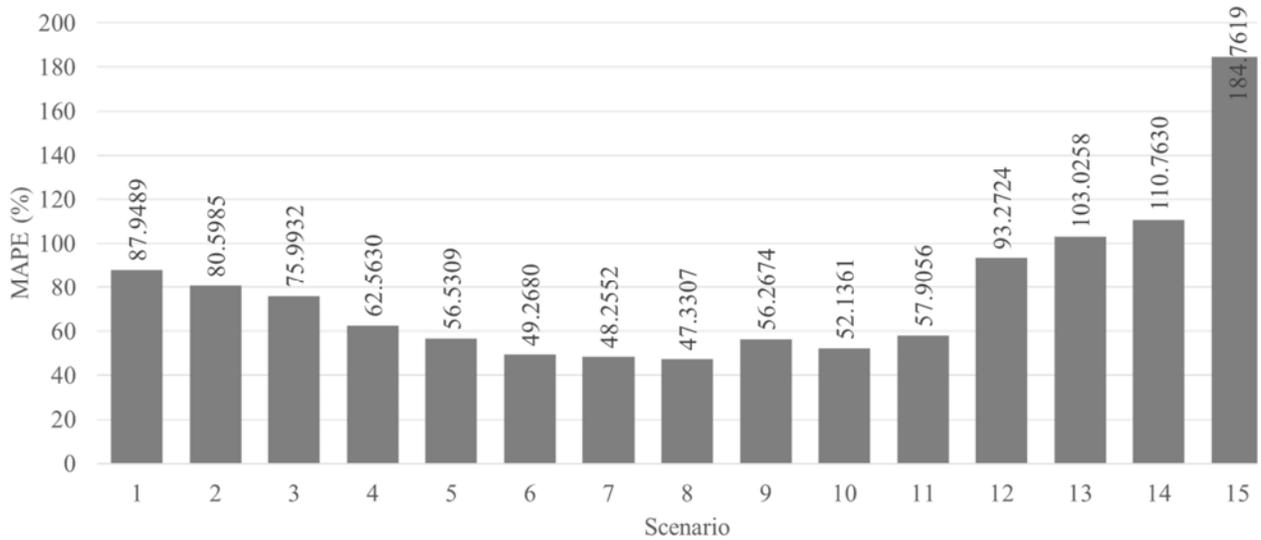

Fig. 3. Performance evaluation on various ELM as baseline regressors

During data cleaning process, only 9 attributes left to be considered as random variables with 1 target variable, i.e. traffic flow. Clean dataset is then scaled accordingly such that the values range between 0 and 1. It aims to avoid the huge variance in the dataset.

### B. Baseline Regressor

Before implementing DM-CCRF to the data, the baseline regressor needs to be prepared. In this research, a neural-network-based regressor is chosen, i.e. Extreme Learning Machine (ELM) [9]. Several variations of ELM parameters are adjusted to generate various quality of baseline regressors. Hence, the behavior of DM-CCRF when reacts to different quality of baseline can be observed. Table 1 shows the parameter variation of ELM as baseline for DM-CCRF.

Each of scenarios is evaluated in terms of Mean Absolute Percentage Error (MAPE), given as (16) to be the benchmark for DM-CCRF. DM-CCRF is expected to have a better performance compared to ELM's performance. The result shown in Figure 3.

$$\text{MAPE} = \frac{100}{N} \sum_{i=1}^{N} \left| \frac{y - \hat{y}}{y} \right| \qquad (16)$$

## IV. RESULT AND DISCUSSION

DM-CCRF is compared to its standard version using the same baseline and implemented in the same dataset. The results are plotted as in the following Figure 4.

According to Figure 4, there is a significant improvement made by both CCRF and DM-CCRF from the baseline performance. It shows the capability of CCRF approach to decrease the error rate of the baseline. DM-CCRF even can overcome the performance of the standard CCRF. Table II shows head-to-head comparison between DM-CCRF and CCRF that also compared to ELM.

As we can see in Table II, in all scenarios, DM-CCRF always win the head-to-head comparison by showing lower error rate. On the other hand, the standard CCRF seems cannot significantly decline the error rate of its baseline. CCRF cab only decline ELM's error by 0.225% to 4.582% while DM-CCRF performs much better by showing higher difference to ELM's error, i.e. from 5.112% to 9.624%. DM-CCRF shows its best performance in the last scenario by decreasing ELM's error rate from 184.762% to 167.715% while the standard CCRF can only decrease it to 177.132%. It is 9.417% worse than DM-CCRF.

TABLE II. HEAD-TO-HEAD COMPARISON BETWEEN ELM, CCRF, AND DM-CCRF

| Scenario | MAPE (%) | | |
|---|---|---|---|
| | ELM | CCRF | DM-CCRF |
| 1 | 87.949 | 87.112 | **80.312** |
| 2 | 80.598 | 79.521 | **73.916** |
| 3 | 75.993 | 74.774 | **69.706** |
| 4 | 62.563 | 62.281 | **57.903** |
| 5 | 56.531 | 56.404 | **52.663** |
| 6 | 49.268 | 47.667 | **46.314** |
| 7 | 48.255 | 47.328 | **45.342** |
| 8 | 47.331 | 46.265 | **44.966** |
| 9 | 56.267 | 54.286 | **52.796** |
| 10 | 52.136 | 49.747 | **48.400** |
| 11 | 57.906 | 57.067 | **53.297** |
| 12 | 93.272 | 92.585 | **84.893** |
| 13 | 103.026 | 102.459 | **93.925** |
| 14 | 110.763 | 109.814 | **100.349** |
| 15 | 184.762 | 177.132 | **167.715** |
| Average | 77.775 | 76.296 | **71.500** |
| Head-to-head | 0 | 0 | **15** |

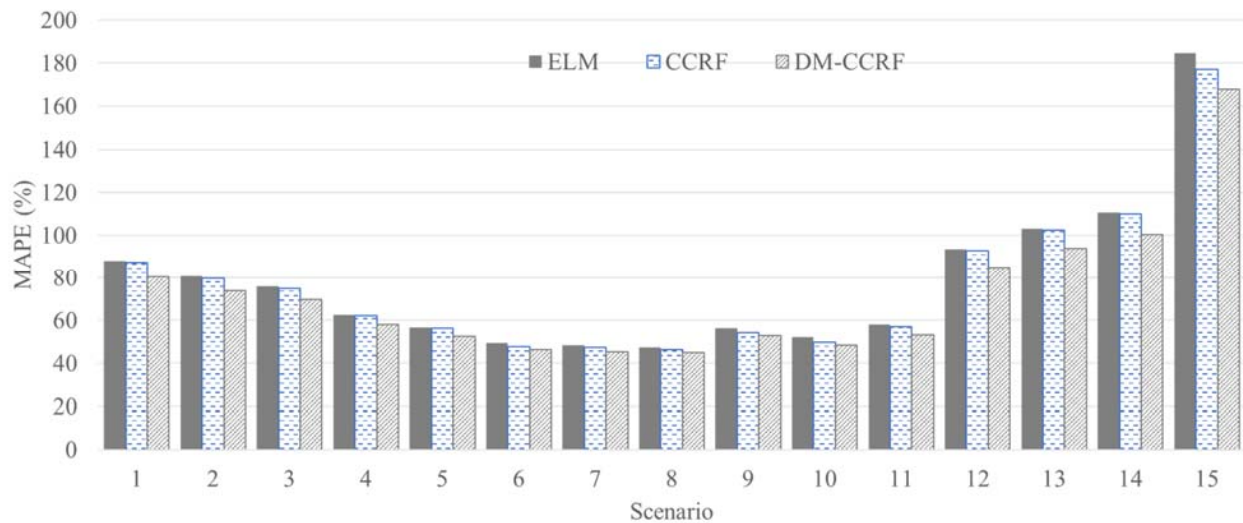

Fig. 4. Performance of DM-CCRF compared to standard CCRF and ELM as its baseline regressor.

## V. Conclusion

In this research, Distance-to-Mean Continuous Conditional Random Field (DM-CCRF) is conducted to predict the highways traffic flow in Highways Agency Dataset. DM-CCRF performance is compared to the standard version of CCRF and to ELM as its baseline regressor. Based on the experiment, it can be concluded that DM-CCRF shows a quite better performance than that of standard CCRF. DM-CCRF contributes a good relationship and continuity between prediction results in this time series dataset. This is validated by the error decrease of ELM up to 9% significance. This is twice the standard CCRF performance which can only decrease Elm's error by 4.582% at most.

## Acknowledgment *(Heading 5)*

We would like to express our sincere gratitude to the Directorate of Research and Community Service Universitas Indonesia for supporting this research through the PITTA Grant scheme No. sss year 2017.

## References

[1] T. R. Board, *Expanding Metropolitan Highways: Implications for Air Quality and Energy Use -- Special Report 245*. Washington DC: The National Academies Press, 1995.

[2] A. Wibisono, W. Jatmiko, H. A. Wisesa, B. Hardjono, dan P. Mursanto, "Traffic big data prediction and visualization using Fast Incremental Model Trees-Drift Detection (FIMT-DD)," *Knowledge-Based Syst.*, vol. 93, hal. 33–46, 2016.

[3] V. Radosavljevic, S. Vucetic, dan Z. Obradovic, "Gaussian Conditional Random Fields for Regression in Remote Sensing," in *ECAI*, 2010, hal. 809–814.

[4] T. Baltrusaitis, N. Banda, dan P. Robinson, "Dimensional Affect Recognition using Continuous Conditional Random," in *IEEE International Conference on Automatic Face and Gesture Recognition (FG)*, 2013.

[5] T. Baltrusaitis, "Automatic facial expression analysis," University of Cambridge, 2014.

[6] T. Baltrusaitis, P. Robinson, dan L. Morency, "Continuous Conditional Neural Fields for Structured Regression," no. 1, hal. 1–16.

[7] V. Imbrasaite, T. Baltrusaitis, dan P. Robinson, "CCNF for Continuous Emotion Tracking in Music: Comparison with CCRF and Relative Feature Representation," in *Multimedia and Expo Workshops (ICMEW), 2014 IEEE International Conference on*, 2014, hal. 1–6.

[8] T. Qin, T. Liu, X. Zhang, D. Wang, dan H. Li, "Global ranking using continuous conditional random fields," *Adv. Neural Inf. Process. Syst.*, hal. 1281–1288, 2009.

[9] G. Huang, Q. Zhu, dan C. Siew, "Extreme learning machine : Theory and applications," *Neurocomputing*, vol. 70, hal. 489–501, 2006.